\documentclass[10pt,twocolumn,a4paper]{article}

\usepackage[final,pagenumbers]{cvww}

\usepackage{amsmath}
\usepackage{comment}
\usepackage{graphicx}
\usepackage{siunitx}
\usepackage{stmaryrd}
\usepackage[export]{adjustbox}
\usepackage{pgfplots}
\usepackage{xcolor}

\definecolor{LightGray}{RGB}{150, 150, 150}
\definecolor{DarkGreen}{RGB}{0, 162, 42}

\usepackage[pagebackref,breaklinks,colorlinks,allcolors=cvwwblue]{hyperref}

\def\paperID{29}

\title{Human Pose-Constrained UV Map Estimation}

\author{Matej Suchanek, Miroslav Purkrabek, Jiri Matas
\vspace{2mm}\\
Visual Recognition Group\\
Department of Cybernetics\\
Faculty of Electrical Engineering\\
Czech Technical University in Prague\\
{\tt\small sucham11@fel.cvut.cz}
}

\begin{document}
\maketitle



\begin{abstract}
    UV map estimation is used in computer vision for detailed analysis of human posture or activity.
    Previous methods assign pixels to body model vertices by comparing pixel descriptors independently, without enforcing global coherence or plausibility in the UV map.
    We propose Pose-Constrained Continuous Surface Embeddings (PC-CSE), which integrates estimated 2D human pose into the pixel-to-vertex assignment process.
    The pose provides global anatomical constraints, ensuring that UV maps remain coherent while preserving local precision.
    Evaluation on DensePose COCO demonstrates consistent improvement, regardless of the chosen 2D human pose model.
    Whole-body poses offer better constraints by incorporating additional details about the hands and feet.
    Conditioning UV maps with human pose reduces invalid mappings and enhances anatomical plausibility.
    In addition, we highlight inconsistencies in the ground-truth annotations.
\end{abstract}

\section{Introduction}
\label{sec:intro}

Analysis of human pose is an essential part of many computer vision problems and is used in a number of applications, including recognition of human activity, gestures and interaction, detection of people and their intent in autonomous driving scenarios, etc.

Information about the human body can be estimated at different levels of resolution. The simplest is the detection of a bounding box that surrounds the person depicted. This can be more precisely delineated by body segmentation. \emph{Pose estimation}, which estimates the locations of some body keypoints, provides another level of granularity. The most detailed is provided by \emph{UV map estimation} (UVME), where every image pixel is mapped to the surface of a generalized human body. The surface is represented as a mesh with a fixed set of vertices.

\begin{figure}[tb]
    \centering
    \includegraphics[width=0.495\linewidth,page=2]{imgs/examples/UVMap_visualizations.pdf}
    \hfill
    \includegraphics[width=0.495\linewidth]{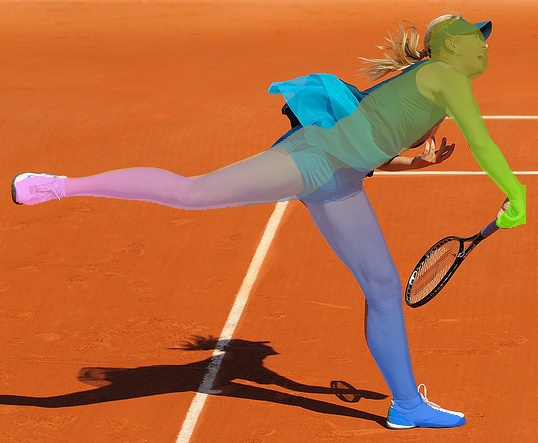}
    \caption{
    The Continuous Surface Embedding method (CSE) \cite{DensePoseCSE} (left) vs. Pose-Constrained CSE (right).
    The CSE method assigns each pixel of body segmentation to a vertex, and thus UV coordinate, on a canonical body shape mesh.
    The CSE assigns each pixel independently, leading to artifacts such as limb duplication (yellow circles).
    \mbox{PC-CSE} uses pose constraints during UV map estimation, producing smoother maps and eliminating artifacts.
    The UV values at individual pixels are visualized by color coding. The location of a given color on the canonical surface is shown in the inset image at the top left.
    }
    \label{fig:teaser}
\end{figure}

The state-of-the-art methods for these tasks \cite{ViTPose,DensePoseCSE,BodyMap} rely on supervised learning, which possibly requires a large amount of annotated data. The cost and effort to annotate the data for human detection, segmentation, pose estimation, and UV map estimation increases with the complexity of the underlying task. UVME is arguably the most complex of these tasks and, therefore, the most data-hungry.

In a recent paper, a method for UVME called Continuous Surface Embeddings (CSE) was introduced \cite{DensePoseCSE}.
The accuracy of the method is good, but it also has limitations.
Due to the disparity between the resolution of the input image and the relatively small number of vertices, this method cannot perform one-to-one matching.
Since each pixel is mapped independently of the others, the method can assign the same body part to multiple locations in the image or produce undesirable artifacts. Examples can be seen in \cref{fig:teaser} and \ref{fig:examples}.

In this paper, our objective is to leverage the methods for pose estimation, which have been in development for a considerable amount of time, to make UVME more accurate. We take advantage of their robustness and design, which guarantees no duplicate assignments.
We introduce the concept of \emph{pose-induced proximal regions} which constrain the mapping to a particular body part and propagate these constraints to the corresponding pixels.

We present a novel method called Pose-Constrained CSE (\mbox{PC-CSE}) that demonstrates the effectiveness of these concepts.
It makes UV maps more coherent with essentially no loss of efficiency besides the need to calculate the human pose.
PC-CSE shows consistent improvement over unconstrained UV maps.
We conducted a detailed ablation study to justify our design choices and explain the improvement in performance.
\section{Related Work}
\label{sec:related}

Human Pose Estimation (HPE) and UV Map Estimation (UVME) are closely related tasks. UVME provides more detailed and comprehensive information, while HPE benefits from a longer history of research, larger datasets, and greater robustness. In this work, we condition UVME predictions on HPE due to HPE's superior reliability. To establish context, we first discuss related work on HPE before moving to UVME advancements.

\textbf{Data.} Progress in human pose and gesture understanding relies heavily on large-scale datasets. The COCO dataset \cite{COCO}, with over 200,000 annotated images of people, is the most widely used, supporting tasks like object detection, instance segmentation, and pose estimation. Its annotations have been extended to whole-body keypoints \cite{COCOWholeBody} and UV map annotations \cite{DensePose}. Other datasets, such as MPII \cite{MPII}, CrowdPose \cite{CrowdPose}, and OCHuman \cite{OCHuman}, target specific challenges like crowded scenes or people in close proximity. While these datasets have significantly advanced research, there is limited research on their overall annotation quality \cite{COCOerrors}.

Current \textbf{2D Human Pose Estimation (HPE)} methods are categorized into top-down, bottom-up, and hybrid approaches. Top-down methods \cite{ViTPose,RTMPose,HRNet} first detect individuals using off-the-shelf person detectors, followed by pose estimation for each detected instance. ViTPose \cite{ViTPose} represents the state-of-the-art in this category. Bottom-up methods \cite{DEKR,OpenPose,BottomUpSeg} predict all keypoints simultaneously and group them into individual poses, making them more effective in crowded scenarios, such as those encountered in OCHuman \cite{OCHuman}. Hybrid approaches \cite{BUCTD} combine elements of both strategies, striking a balance between accuracy and efficiency under challenging conditions.

\textbf{UV Map Estimation (UVME)} has seen steady progress in recent years.
DenseReg \cite{DenseReg} formulates UVME as a regression task and trains a fully convolutional neural network for human face extraction using facial landmarks. DensePose \cite{DensePose}, a milestone in UVME, collects a dataset of many body-to-surface annotations and adapts the Mask R-CNN architecture \cite{MaskRCNN} for person detection, segmentation and UV map estimation in a cascade. Subsequent works focus on seeking correspondences in sequences of images \cite{SlimDensePose,HumanGPS}, utilize DensePose as an intermediate representation for other advanced tasks, such as 3D body reconstruction \cite{Tex2Shape,MeshPose}, or use it as the ground truth \cite{TikTokDataset}.

DensePose relies on splitting the body template into small partitions (``charts'') and performs a simultaneous regression of the target body part and the UV coordinate within the respective partition. Continuous Surface Embeddings (CSE) \cite{DensePoseCSE} follows up on DensePose by eliminating the need for artificial slicing of the template. Instead, CSE holds trainable descriptors (embeddings) of the template surface and guides a neural network to regress these embeddings per pixel in a contrastive manner. The UV map is determined by finding the closest surface embedding of every pixel. Overall, CSE simplifies the DensePose framework while making it generalizable to other natural objects. Both DensePose and CSE are tightly bound to the mesh of the SMPL \cite{SMPL}, a parametrized 3D model of the human body.

BodyMap \cite{BodyMap} further refines CSE by addressing body details such as hair and clothing, providing high-fidelity results while relying on CSE descriptors internally. Although it claims state-of-the-art performance, its code has not been released to the public. Recently, foundational models like Sapiens \cite{Sapiens} have emerged in human-centric vision tasks. Trained on vast amounts of unannotated data, these models achieve state-of-the-art performance across various downstream tasks. However, they are resource-intensive and have yet to demonstrate significant advancements, specifically in UV map estimation.

\begin{figure*}[tb]
    \centering
    \begin{subfigure}[t]{0.252\linewidth}
        \centering
        \includegraphics[width=\textwidth,height=41ex,valign=t,page=13]{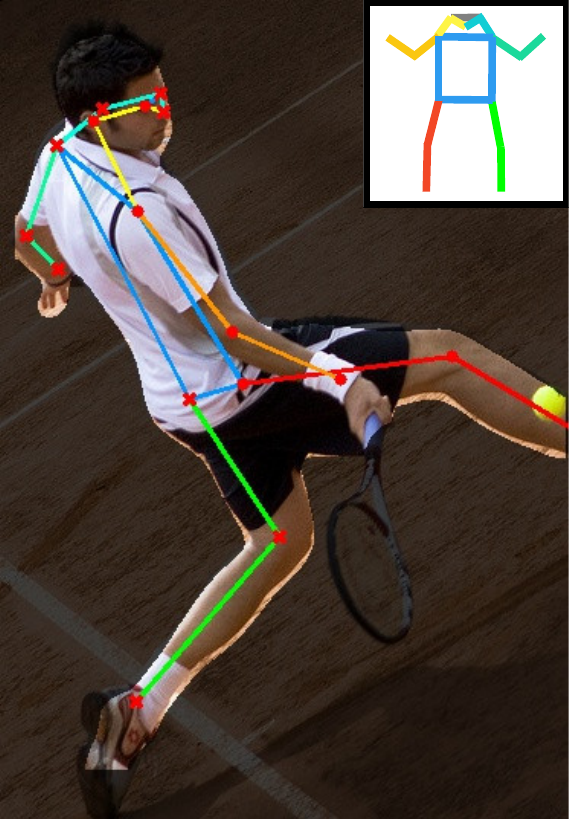}
        \caption{
        PC-CSE requires a bounding box and a segmentation mask as an input.
        \mbox{VitPose-l} \cite{ViTPose} is used for pose estimation in this example.
        Front-view skeleton is in the inset image.
        }
        \label{fig:pipeline-vis-pose}
    \end{subfigure}
    \hfill
    \begin{subfigure}[t]{0.47\linewidth}
        \centering
        \includegraphics[width=\textwidth,height=41ex,valign=t,page=9]{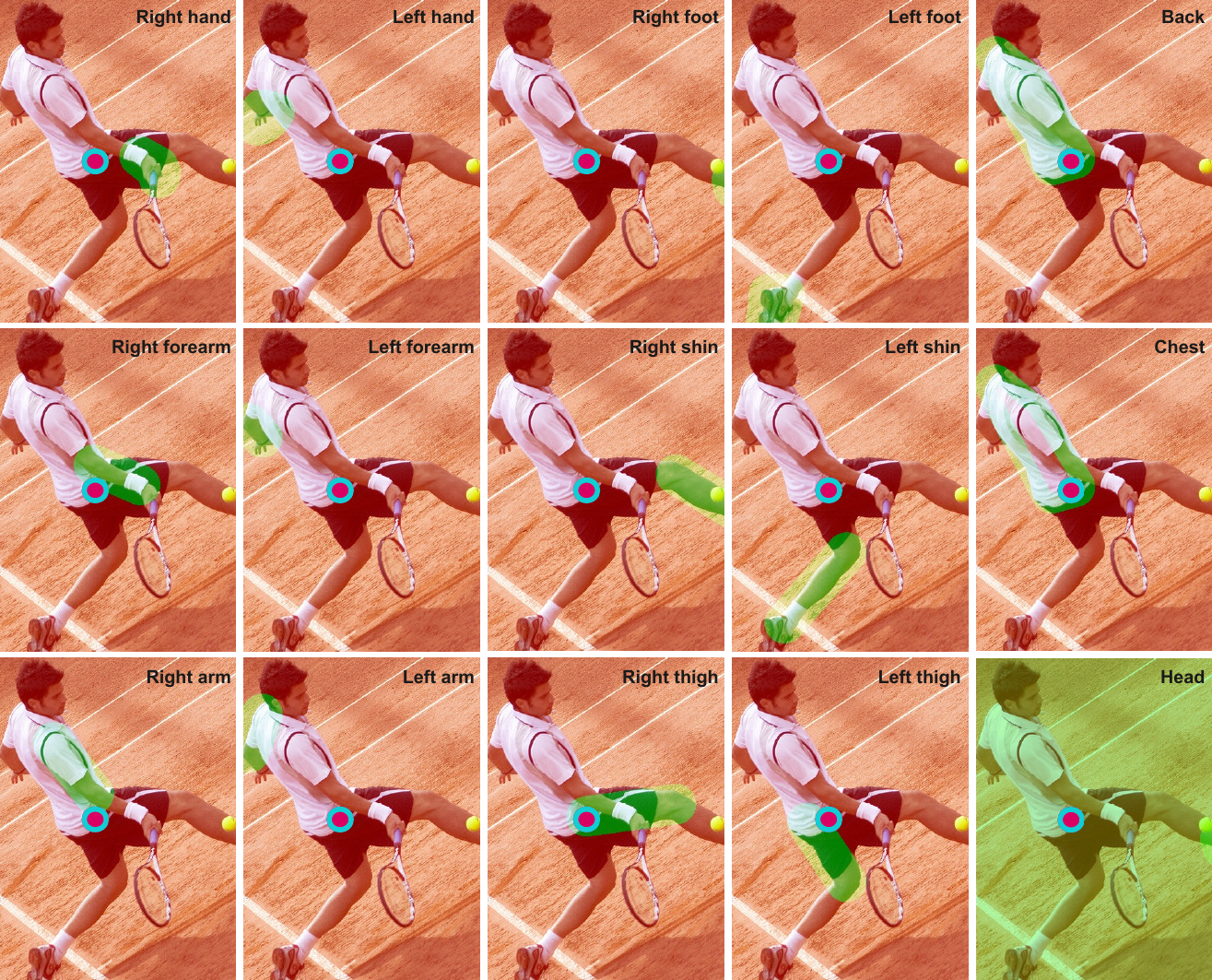}
        \caption{
        Proximal regions of body parts.
        Only pixels in the segmentation mask and the green areas may be assigned to the body parts denoted in the top right. Head and undetected body parts are unconstrained (bottom right).
        The highlighted point can be assigned to the back, chest, head, and both the left and the right thigh but not to other parts.
        }
        \label{fig:pipeline-vis-proximal}
    \end{subfigure}
    \hfill
    \begin{subfigure}[t]{0.252\linewidth}
        \centering
        \includegraphics[width=\textwidth,height=41ex,valign=t,page=8]{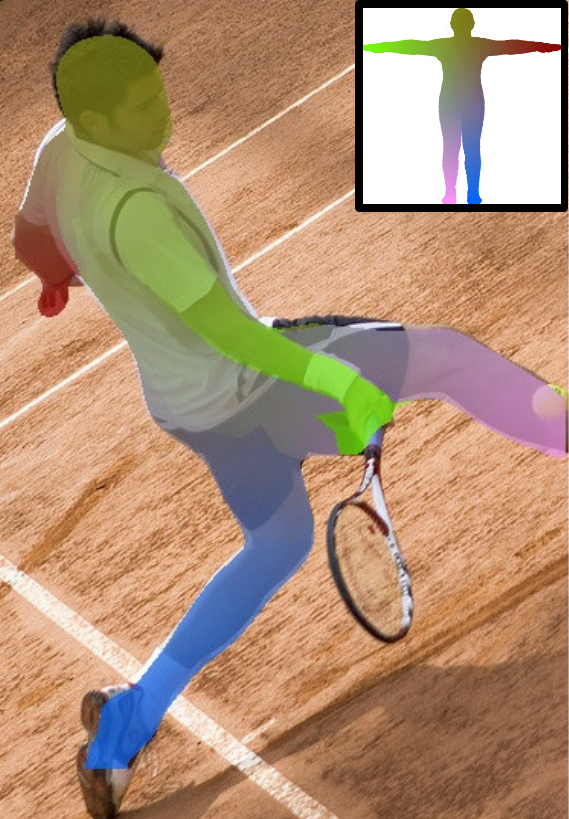}
        \caption{
        The PC-CSE UV map is consistent with the estimated pose, unlike the CSE \cite{DensePoseCSE} estimate.
        The difference is shown in \cref{fig:examples}.
        }
        \label{fig:pipeline-vis-constr}
    \end{subfigure}
    \caption{
    Pose-constrained CSE (\mbox{PC-CSE}) takes an estimated bounding box, segmentation mask, and 2D human pose (a) as input.
    It computes proximal regions (b) for each body part and assigns pixels to SMPL~\cite{SMPL} vertices to generate a UV map.
    Unlike the CSE~\cite{DensePoseCSE}, \mbox{PC-CSE} constrains pixel assignments using proximal regions, ensuring the resulting UV map aligns with the estimated pose (c).
    }
    \label{fig:pipeline-vis}
\end{figure*}

\section{Method}
\label{sec:method}

Our method is built on top of the CSE method \cite{DensePoseCSE}, a feed-forward neural network based on the Mask \mbox{R-CNN} architecture \cite{MaskRCNN}. Although it performs human detection, segmentation and UV map estimation in a cascade, we are concerned only with the latter and consider bounding boxes and segmentation as input determined by an external method.

The network outputs pixel descriptors, or \emph{pixel embeddings}. During training, contrastive learning is employed to determine both the best weights and the values of \emph{vertex embeddings}, each linked to one of the vertices of the SMPL mesh \cite{SMPL}. The resulting UV map is established by mapping every input pixel embedding to the most similar vertex embedding (in terms of cosine similarity), associating every image pixel with a mesh vertex (and its UV coordinates).

Formally, let $I$ be the input image, $x \in I$ a (foreground) image pixel, $\Phi_x(I) \in \mathbb{R}^D$ embedding of the pixel $x$ provided by the neural network $\Phi$ (where $D$ is the embedding dimensionality), $M$ the mesh (set of vertices), $i \in M$ a~vertex index, and $E_i \in \mathbb{R}^D$ normalized embedding of the vertex~$i$. The mapping from pixels to vertices using CSE \cite{DensePoseCSE} can be expressed as:
\begin{equation}
  i^{*}_{x} = \arg\max_{i \in M} \left\langle E_i, \Phi_x(I) \right\rangle.
  \label{eq:CSE}
\end{equation}

Consistent with the standard definition of mapping, CSE always maps exactly one vertex to every foreground pixel. However, the reverse is not necessarily true. Typically, the resolution of the input image is sufficiently high such that the pixel count significantly surpasses the vertex count on the mesh, resulting in multiple pixels being frequently associated with the same vertex.

Arguably, this does not pose a problem in itself. For example, it is acceptable for neighboring pixels to map to the same vertex, as they can lie so close to each other on the actual body that the discretization of the mesh cannot distinguish between them. Nevertheless, a fully independent assignment of vertices to foreground pixels makes CSE generally prone to implausible pose predictions, as it has no other means to avoid them but to rely on the strength of its prior. Qualitative research confirms our hypothesis, as we observe situations stemming from the general problem, such as CSE assigning the same body part to more than one image region (\eg, two hands are declared left), UV map discontinuities and various artifacts (see \cref{fig:examples}).

At this point, we examine the features of human pose estimation (HPE) algorithms. These estimators predict the locations of various landmarks on the human body, called \emph{keypoints}, such as skeletal joints or facial landmarks. In particular, skeletal joints form a primitive human skeleton, the shape of which is very similar to that of our 3D human representation (\cref{fig:pipeline-vis-pose}). Furthermore, each keypoint is, by design, assigned to at most one image coordinate. This constitutes the key advantage of HPE over CSE, as duplicate assignments of body parts become impossible.

\subsection{Conditioning CSE by pose}
\label{sec:method-UVbyPose}

We believe that using a human pose estimation model as a secondary expert during inference and enforcing consistency of the two representations is a promising path for avoiding errors in predicted UV maps and improving their quality.
Therefore, we propose our new method called \textbf{Pose-Constrained Continuous Surface Embeddings} (\mbox{PC-CSE}).
The key enhancement is the introduction of \emph{pose-induced constraints} whose purpose is to limit the mapping of every pixel to only pre-selected body regions.
It does not involve any architectural change to CSE and does not require its retraining or fine-tuning.

The constraints are rules that determine to which vertices of the mesh each foreground pixel is allowed to map.
Which pixels are constrained by which rule depends on the inferred pose.
We first define the relation between the pose representation and the target mesh.
We use the COCO skeleton \cite{COCO} as the default pose representation.
It consists of 17 keypoints (\cref{fig:pipeline-vis-pose}): 12 skeletal joints (wrists, elbows, shoulders, hips, knees, and ankles in pairs) and 5 facial landmarks (eyes, ears, and nose).
These keypoints can be linked into arms, forearms, thighs, shins, and a quadrilateral defined by shoulders and hips.
We refer to these connections as the \emph{principal bones}.

In addition, we explore the \emph{whole-body} skeleton \cite{COCOWholeBody}.
This representation with 133 keypoints extends the COCO skeleton by introducing extra keypoints for hands, feet, and face.
This poses an advantage over the basic version because hands and feet are somewhat distant from the respective keypoints and can deviate from the limb axis.

The canonical mesh can now be partitioned into subsets of vertices.
Each partition should roughly correspond to one principal bone.
We create 15 mesh partitions of SMPL -- arms, forearms, hands, thighs, shins, and feet in pairs, the front and back of the torso, and head -- by merging segments of SMPL body segmentation \cite{SMPL,Meshcapade}.
We divide the torso by the sagittal plane to distinguish between the front and back of it.

The scope of constraints within the image is specified by expanding (``inflating'') the inferred skeleton composed of the principal bones.
Each principal bone delineates its \emph{proximal region}, each defined as a set of pixels with a certain maximum pixel distance from the bone (\cref{fig:pipeline-vis-proximal}).
The optimal distance obviously relies on the apparent size of the person (which varies with its distance from the camera) and needs to be determined for each person separately.
We try to estimate it using an algorithm that also depends on the pose; it is described in detail in \cref{sec:method-areas}.

The capsular shape of the proximal regions is most appropriate for the limbs, \ie, arms, forearms, thighs, and shins.
Concerning the front and back of the torso, we first merge the central quadrilateral (\ie, between the shoulders and hips) with the regions around its sides, which we also define as having a capsule-like shape.
Then, we analyze the mutual position of its corners to discriminate between the frontal and dorsal view.
If the orientation of the keypoints implies the frontal view of the person, we subtract the quadrilateral from the back, and vice versa (see the rightmost column of \cref{fig:pipeline-vis-proximal}).

Nonetheless, the basic COCO skeleton does not adequately support precise localization of the hands (fingers) and feet (toes).
Various strategies can be employed to manage this.
With the whole-body skeleton, the proximal regions for these body parts can span the extra keypoints.
As a fallback when using the basic skeleton, we propose circular proximal regions around the closest keypoint (wrist for hands, ankle for feet) twice as wide as the capsular ones.
Both these options are discussed in the experiments (\cref{sec:experiments}).
Otherwise, a conservative approach is to merge body parts with the nearest bone or leave them unconstrained, but this does not fully leverage the capabilities of our method.
In addition, we do not outline a dedicated proximal region for the head, but we let all pixels map to it.
We consider the head to be easily recognizable, and our primary goal is to resolve duplications between paired limbs.

The proximal regions induce semantic labeling of image pixels by template partitions.
Every pixel is labeled according to the proximal regions to which it belongs.
If multiple proximal regions overlap, the pixels within the intersection are labeled with all corresponding labels.
If a pixel falls outside all proximal regions, it gets all possible labels (thus, it keeps the original prediction).
When a body part is missing (that is, either of its keypoints is not provided by the HPE model), we allow mapping to it from any foreground pixel.
The purpose of this rule is to prevent inaccurate refinements where, for example, a forearm is partially visible, but one of its ends lies outside the image.
As described earlier, we always apply this rule to the head as well.

As a result, each pixel receives information about its target body part(s) implied by the pose-induced constraints and the embedding provided by the original CSE.
We now modify the original procedure (\cref{eq:CSE}) to consider the constraints as well.
Instead of yielding the vertex with the highest similarity of all mesh vertices, we limit the output space to one of those vertices that belong to the body partitions defined by the constraints.
The chosen vertex (its embedding) should still have the highest similarity to the pixel embedding, but only vertices from the limited subset of the whole mesh should be considered.

Formally, let $p \in P$ be the partition label (index), $M_p \subset M$ the vertices of the partition $p$, $L \colon I \to \mathcal{P}(P) \setminus \{\emptyset\}$ a function mapping a pixel to a set of allowed partitions.
\Cref{eq:CSE} now becomes:
\begin{equation}
  i^{*}_{x} = \arg\max_{i \in M_{L(x)}} \left\langle E_i, \Phi_x(I) \right\rangle,
  \label{eq:PC-CSE}
\end{equation}
where
\begin{equation}
  M_{L(x)} = \bigcup_{p \in L(x)} M_p.
  \label{eq:union}
\end{equation}

Alternatively, let $B(x, p)$ be the binary flag (0 or 1) indicating whether partition $p$ is allowed in pixel $x$, $V(x, p)$ the vertex from partition $p$ with the highest similarity to pixel $x$, $S(x, p)$ the similarity of vertex $V(x, p)$ to pixel $x$ and $S'(x, p)$ our adjusted similarity. We compute these matrices as follows:
\begin{align}
    B(x, p) &= \llbracket p \in L(x) \rrbracket, \\
    V(x, p) &= \arg\max_{i \in M_p} \left\langle E_i, \Phi_x(I) \right\rangle, \\
    S(x, p) &= \max_{i \in M_p} \left\langle E_i, \Phi_x(I) \right\rangle, \\
    S' &= S \odot B.
\end{align}
\Cref{eq:PC-CSE} is then equivalent to:
\begin{equation}
    i^{*}_{x} = V(x, \arg\max_{p \in P} S'(x, p)).
\end{equation}
We believe that this approach is more practical for implementation as it avoids computing unions of mesh partitions (\cref{eq:union}) and storing them in memory.

\subsection{Determining proximal regions}
\label{sec:method-areas}

As an intermediate step, \mbox{PC-CSE} expands the inferred human skeleton so that its shape approximately matches the silhouette (segmentation) of the person. The exact expansion range is a trade-off. Small proximal regions might not adjust the UV map at full width. Large proximal regions can cause significant overlaps with each other, making pose-induced constraints less effective. In extreme cases, the expansion range can be chosen as zero, resulting in no correction made, or it can be chosen so high that every proximal region covers the whole body. We note that in both cases, the new prediction would be the same as, and thus \emph{not worse than}, the original prediction.

The expansion range should roughly correspond to the width (thickness) of the person's limbs, expressed in pixel units. We further refer to it as the \emph{bone width} ($\Delta$) and assume that it is proportional to other measures of the body, in particular the person's height. Typically, information about body measures is accessible only in controlled environments where the camera model and relative location of the object and camera are known. However, this requirement would significantly limit our method and render it useless for data ``in the wild''.

Thus, we introduce a technique for estimating these measures based only on information about the person's pose. The prerequisite is knowledge of the actual (3D) lengths of the principal bones determined by the pose estimation model. We obtain these distances from SMPL \cite{SMPL}. During inference, we measure the distances in the pixel space and normalize (divide) them by their distance in the 3D space. Each measurement serves as an estimate of one SMPL model unit length in pixels, assuming that the bone is parallel to the projection plane.

We then apply simple trigonometry-based reasoning to choose the most credible estimate. Given a straight unit-length stick parallel to the ground plane, its apparent length is maximal when it is parallel to the projection plane, too, and decreases when rotating the stick around the vertical axis (down to zero when both ends visually merge to the same point). In our domain, sticks are the principal bones of different lengths. Normalizing the distances by the respective lengths makes the estimates proportional only to the cosine of the angle with the projective plane. Since cosine is a decreasing function of angle (for $\alpha \in [\ang{0}, \ang{90}]$), the bone having the smallest angle (ideally zero) with the projection plane will correspond to the highest value. Therefore, the best estimate is the \emph{maximum}.

Arguably, this estimate cannot be considered perfect since we have no guarantee that the assumption of parallelism actually holds. However, we are interested in determining the size of proximal regions, which do not need to match the shape of the person exactly. In fact, a minor overestimation of the size is not a problem because we do not deal with pixels in the background anyway, and it can also help us handle people with different body mass. 

Therefore, we determine the best multiplication factor by tuning it using the validation data. The results are presented in the ablation study (\cref{sec:ablation}).
\section{Data}
\label{sec:data}

\begin{figure}[tb!]
    \centering
    \begin{subfigure}{\linewidth}
        \centering
        \includegraphics[width=0.495\textwidth,height=0.68\textwidth,page=1]{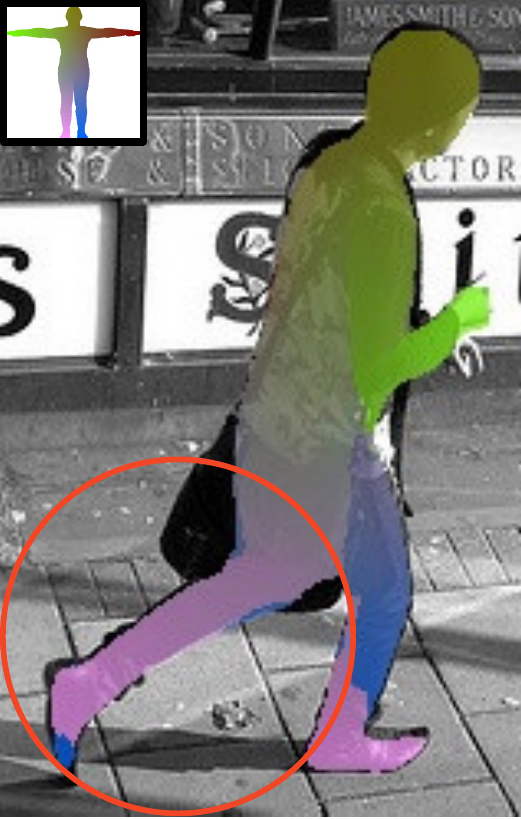}
        \hfill
        \includegraphics[width=0.495\textwidth,height=0.68\textwidth]{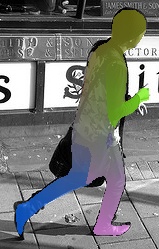}
    \end{subfigure}
    \begin{subfigure}{\linewidth}
        \centering
        \includegraphics[width=0.495\textwidth,page=6,height=0.40\textwidth]{imgs/examples/UVMap_visualizations.pdf}
        \hfill
        \includegraphics[width=0.495\textwidth,height=0.40\textwidth]{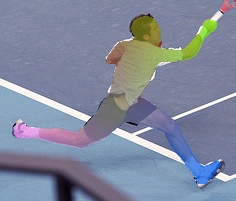}
    \end{subfigure}
    \begin{subfigure}{\linewidth}
        \centering
        \includegraphics[width=0.495\textwidth,page=3]{imgs/examples/UVMap_visualizations.pdf}
        \hfill
        \includegraphics[width=0.495\textwidth]{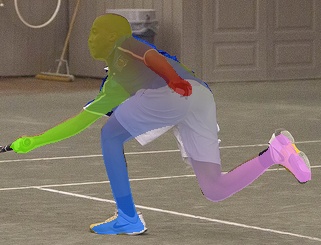}
    \end{subfigure}
    \begin{subfigure}{\linewidth}
        \centering
        \includegraphics[width=0.495\textwidth,height=0.65\textwidth,page=4]{imgs/examples/UVMap_visualizations.pdf}
        \hfill
        \includegraphics[width=0.495\textwidth,height=0.65\textwidth]{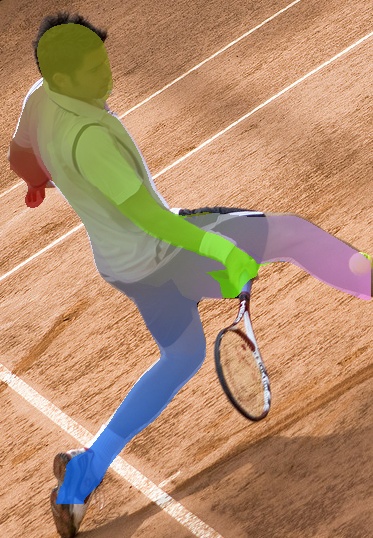}
    \end{subfigure}
    \begin{subfigure}{\linewidth}
        \centering
        \includegraphics[width=0.495\textwidth,page=5]{imgs/examples/UVMap_visualizations.pdf}
        \hfill
        \includegraphics[width=0.495\textwidth]{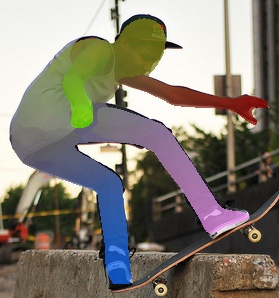}
    \end{subfigure}
    \caption{
    CSE \cite{DensePoseCSE} (left) vs. \mbox{PC-CSE} conditioned by estimated pose (right).
    Pose constraints ensure smoother UV maps and prevent limb duplication within a single image. A frontal view of the SMPL model \cite{SMPL} is shown to help assess the UV~estimation. 
    }
    \label{fig:examples}
\end{figure}

In our experiments, we rely on the DensePose COCO dataset \cite{DensePose}. This dataset contains about 50 thousand annotated people on a subset of images from the COCO dataset \cite{COCO}. In addition to the bounding box coordinates, instance segmentation mask, and keypoints (skeleton), the ground-truth information about every instance includes the body segmentation mask and a set of dense correspondences (over 5 million annotated points in total).

The dataset is divided into train and validation splits with a ratio of about 95/5.

\subsection{Assessing the quality of annotations}
\label{sec:quality}

During our research, we repeatedly encountered incorrectly annotated instances in DensePose COCO.
Therefore, as part of our efforts, we conducted research on their overall quality.
We define miscellaneous metrics that express the consistency of an instance's ground truth data.
(For more details, see supplementary.)
Then, we manually inspect the lowest-ranking instances and identify the most common problems:

\begin{enumerate}
    \item Annotators of dense correspondences confuse the left and right parts of the body. In most cases, only one pair of body parts is confused, while the rest are annotated correctly.
    \item Dense correspondences of thighs and shins are even more confused. Some instances are annotated as having only the left or only the right leg, or annotations of one leg have mixed laterality.
    \item Keypoint annotators more often confuse left and right per limb or the orientation of the entire body rather than a single pair of keypoints.
    \item When multiple people at least partially overlap with a bounding box, the annotated instance is different from the one that matches the dimensions of the bounding box.
    \item Body segmentation masks are incomplete; not all body parts are segmented.
    \item Bounding boxes lack the ``is crowd'' label. These are supposed to annotate many people at once (\ie, a crowd) and should not be associated with dense or keypoint annotations.
\end{enumerate}

We do not make any corrections to the ground truth, but we remove dense annotations that we consider wrong. We assess the precision per body part, not individually per point. If a body part shows any of the above problems, we remove all associated points regardless of laterality. As a result, we remove ca.~1.5\% points from the dataset, concerning ca.~7.5\% instances. (For the validation subset, the numbers are somewhat higher: 2.4\% points on 11.2\% instances.)

\section{Experiments}
\label{sec:experiments}

In the following, we evaluate \mbox{PC-CSE} by simulating its use in practice.
We take the \texttt{R\_101\_FPN\_DL\_soft\_s1x} CSE model from the \mbox{detectron2} toolbox \cite{detectron2} and consider it to be the baseline method.
We run inference on images from the validation subset of the DensePose COCO dataset (\cref{sec:data}) and obtain the baseline bounding boxes, instance segmentation, and pixel embeddings.

Then, we use the bounding boxes as input for top-down HPE models, which we obtain from the \mbox{mmpose} toolbox \cite{mmpose}.
We choose several HPE models that differ in performance and provide different representations of human pose (see \cref{sec:method-UVbyPose}).
Finally, we combine all outputs and apply our \mbox{PC-CSE} method and compare the accuracy of the newly produced UV maps to that of the baseline ones.

\subsection{Evaluation metrics}
\label{sec:metrics}

We follow the modified COCO challenge protocol \cite{COCO} that evaluates the match between predictions and ground-truth instances using Geodesic Point Similarity (GPS) and computes the algorithm's Average Precision (AP) by thresholding the GPS score \cite{DensePose}.
We report the Average Precision for both the original dataset and the dataset without incorrect annotations (see \cref{sec:quality}).

\subsection{Results}
\label{sec:results}

\Cref{tab:results} compares pose-constrained CSE (PC-CSE) with the original CSE \cite{DensePoseCSE}.
The results are reported in the COCO val dataset for comparability with previous work.
Furthermore, we evaluated performance on the COCO val data set while ignoring incorrect annotations, as described in \cref{sec:quality}.

\begin{table}
   \centering
   \begin{tabular}{@{}l|rrr@{}l}
   \toprule
   HPE method & HPE & UV map & UV map & \makebox[-0pt]{}{$^\dagger$} \\
   \hline
   \textit{None}               & ---    & 66.2  & 68.8  \\
   
   \mbox{ViTPose-b} \cite{ViTPose}    & 75.8   & 66.8  & 69.3  \\
   \mbox{ViTPose-h} \cite{ViTPose}    & 79.1   & 67.0  & 69.6  \\
   \mbox{ViTPose-h wb}                & 78.6   & 67.3  & 69.8  \\
   \hline
   \mbox{RTMPose-l} \cite{RTMPose}    & 75.8  & 67.0  & 69.5  \\
   \mbox{RTMPose-l wb} \cite{RTMPose} & 69.5  & 66.7  & 69.3  \\
   
   \bottomrule
   \end{tabular}
   \caption{
    \textbf{AP results on the COCO dataset}.
    Constraining UV map estimation with 2D pose improves performance.
    More accurate poses lead to better UV maps.
    Using the whole-body (wb) skeleton further enhances performance due to better hand and foot constraints.
    Note that 2D Human Pose Estimation (HPE) is evaluated on a different COCO subset than UV map evaluation.
    Results marked with ($^{\dagger}$) are evaluated on data with ignored incorrect annotations, as detailed in \cref{sec:quality}.
   }
   \label{tab:results}
\end{table}

\begin{figure}[tb]
    \centering
    \begin{tikzpicture}
    \begin{axis}[
        width=\linewidth,
        title={},
        xlabel={Bone width $\Delta$},
        ylabel={COCO val AP},
        xmin=0, xmax=1.25,
        ymin=68.72, ymax=69.78,
        xtick={0.08, 0.40, 0.60, 0.8, 1.0, 1.2},
        ytick={68.25, 68.5, 68.75, 69.0, 69.25, 69.50, 69.75, 70.0},
        legend pos=north east,
        ymajorgrids=true,
        grid style=dashed,
    ]

    \addplot[
        color=red,
        dashed,
        line width=1.2pt,
        mark=circle,
        ]
        coordinates {
        (0.00,68.8)
        (1.25,68.8)
        };
    \addplot[
        color=blue,
        mark=*,
        line width=1.2pt,
        ]
        coordinates {
        (0.06,69.2)
        (0.07,69.4)
        (0.08,69.6)
        (0.10,69.5)
        (0.15,69.4)
        (0.20,69.3)
        (0.50,69.0)
        (1.00,68.8)
        (1.20,68.8)
        };
    \addplot[
        color=black,
        dotted,
        line width=1.2pt,
    ]
    coordinates {
        (0.085,69.6)
        (0.085,68.5)
    };
    \legend{
    No constraints,
    PC-CSE
    }

    \end{axis}
    \end{tikzpicture}
    \caption{
    \textbf{Ablation on bone width $\Delta$} defined in \cref{sec:method-areas}.
    \mbox{RTMPose-l wb} \cite{RTMPose} is used for pose constraints.
    Too thin bones restrict UV Map too much and hinder performance on border pixels.
    Excessively thick bone estimates do not restrict UV Map sufficiently and reduce the performance gain.
    Note that performance with proximal regions with large regions $\Delta$ converges to the baseline method.
    In the extreme case when all bones are as big as the whole picture, no constraints are applied.
    The best value is 0.08.
    }
    \label{fig:ablation-width}
\end{figure}
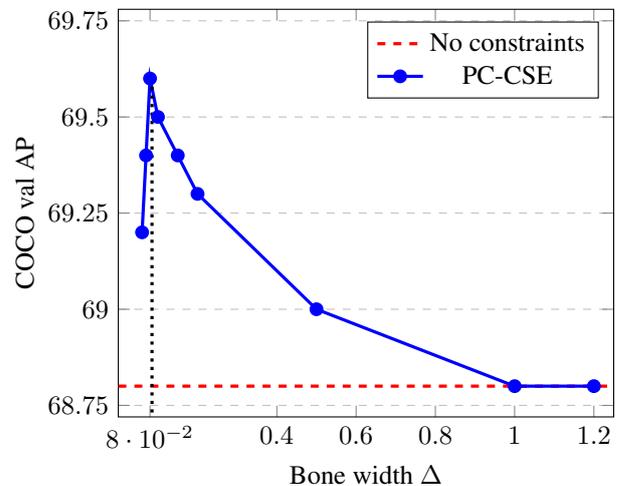

The first row of \cref{tab:results} shows the performance of CSE \cite{DensePoseCSE} without pose constraints.
We reproduced these results and observed a 2.6~AP improvement when ignoring incorrect annotations.
This gain remains consistent across all experiments.

Subsequent rows show results with pose constraints from ViTPose \cite{ViTPose} and RTMPose \cite{RTMPose}, using different model variants.
Regardless of the HPE model, applying pose-conditioned constraints consistently improves performance.
As expected, the performance gain depends on the quality of the HPE model. \mbox{ViTPose-h} (huge) outperforms \mbox{ViTPose-b} (base) in HPE and achieves slightly better UV map accuracy.
However, the difference is minor. Note that HPE is evaluated on a larger subset of COCO images than UV maps.

To assess the impact of the whole-body (wb) skeleton, we trained \textit{ViTPose-h wb} on the COCO-WholeBody dataset \cite{COCOWholeBody}.
It achieves 67.3~AP on COCO-WholeBody and 78.6~AP on COCO, compared to 79.1~AP for \mbox{ViTPose-h}.
While the whole-body poses are less accurate, the inclusion of fingers and toes compensates for this in specific body regions.

Results for RTMPose \cite{RTMPose} follow a similar trend.
Using estimated poses improves the performance of the UV map between models, although exact gains differ.
For instance, \mbox{RTMPose-l} matches \mbox{ViTPose-b} in HPE performance, but achieves slightly higher UV map accuracy.
However, this difference is negligible.

\mbox{RTMPose-l wb} shows a much weaker HPE performance but comparable UV map accuracy.
Although the inclusion of fingers and toes benefits the hand and foot regions, the reduced accuracy of other keypoints diminishes overall gains, making the trade-off less favorable.

While conditioning UV map predictions on pose significantly improves consistency, this translates to only a modest 1~AP point increase in overall performance due to several factors.
The most significant issue is segmentation errors — pixels outside the segmentation mask are not assigned UV map estimates, leading to penalties.
An example is shown in image \cref{fig:errors-analysis}.
Detection errors also impact performance; if a person is not detected, no UV estimation can be performed.

Achieving 100~AP is challenging due to the limitations of ground truth annotations, which are human estimates often obscured by clothing.
In images with loose clothing, these annotations can be highly imprecise, making it difficult to determine whether discrepancies stem from ground truth errors or model predictions.
As a result, images with GPS around 80 already represent strong estimates, as shown in \cref{fig:errors-analysis}.

Examples of significant improvements over the baseline are shown in \cref{fig:teaser} and \cref{fig:examples}.
These include artifact removal, better continuity between limbs, and elimination of redundant body part assignments in baseline UV maps.


\subsection{Ablation study}
\label{sec:ablation}


\begin{figure}
    \centering
    \begin{tikzpicture}
    \begin{axis}[
        width=\linewidth,
        height=0.8\linewidth,
        xlabel={Difference (GPS)},
        ylabel={Frequency},
        xtick align=outside,
        xtick distance=0.1,
        minor tick num=4,
        ymajorgrids=true,
        ymode=log,
        ymin=1,
        ylabel near ticks,
        xlabel style={
            at={(axis description cs:0.5,-0.05)},
            anchor=north
        },
        log origin=infty,
        grid style=dashed,
        bar width=3,
        enlargelimits=true
    ]

    \addplot[
        ybar,
        fill=blue
    ]
    coordinates {
        (-0.14, 1)
        (-0.1, 1)
        (-0.06, 1)
        (-0.04, 6)
        (-0.02, 70)
        (0.0, 1966)
        (0.02, 86)
        (0.04, 37)
        (0.06, 17)
        (0.08, 13)
        (0.1, 8)
        (0.12, 4)
        (0.14, 4)
        (0.16, 3)
        (0.18, 1)
        (0.2, 1)
        (0.24, 2)
        (0.26, 1)
        (0.28, 1)
        (0.5, 3)
        (0.52, 3)
        (0.54, 2)
        (0.56, 1)
        (0.58, 2)
        (0.62, 1)
        (0.68, 1)
    };
    \end{axis}
    \end{tikzpicture}

    \caption{
    Image-wise performance difference of baseline \mbox{CSE \cite{DensePoseCSE}} and \mbox{PC-CSE} with poses from \mbox{ViTPose-h wb}.
    Performance is measured by GPS, frequency is in log scale.
    Positive means better performance of \mbox{PC-CSE}.
    \mbox{PC-CSE} causes only a few performance drops while improving more other cases, some of them dramatically, and keeping the rest about the same.
    }
    \label{fig:histogram}
\end{figure}

The efficiency of \mbox{PC-CSE} depends on a proper outline of the proximal regions, as described in \cref{sec:method-areas}. To ensure overall robustness, we determine the best value of the bone width hyperparameter $\Delta$ by validation. We use the \mbox{RTMPose-l wb} model \cite{RTMPose} and repeat the same experiment while varying the value of the hyperparameter. Note that we use the clean validation dataset that does not contain the incorrect annotations identified (\cref{sec:quality}).

The results, shown in \cref{fig:ablation-width}, confirm our expectations. With an increasing value of the hyperparameter, the precision increases and reaches the maximum when it is equal to 0.08. Increasing it further, we observe a gradual decrease in precision down to the baseline. This supports our earlier statement (\cref{sec:method-areas}) about the best value being a compromise and the consequences arising from a suboptimal choice. Extremely small and large values do not give our method the opportunity to have the desired impact.

Note that our experiments generally assume that the method for estimating a person's measures (\cref{sec:method-areas}) from their pose is accurate. We do not conduct any quantitative experiments on this matter, but we attempt to verify it using qualitative analysis (see supplementary material). 

\begin{figure}
    \centering
    \begin{subfigure}{0.620\linewidth}
        \includegraphics[width=\linewidth,page=11]{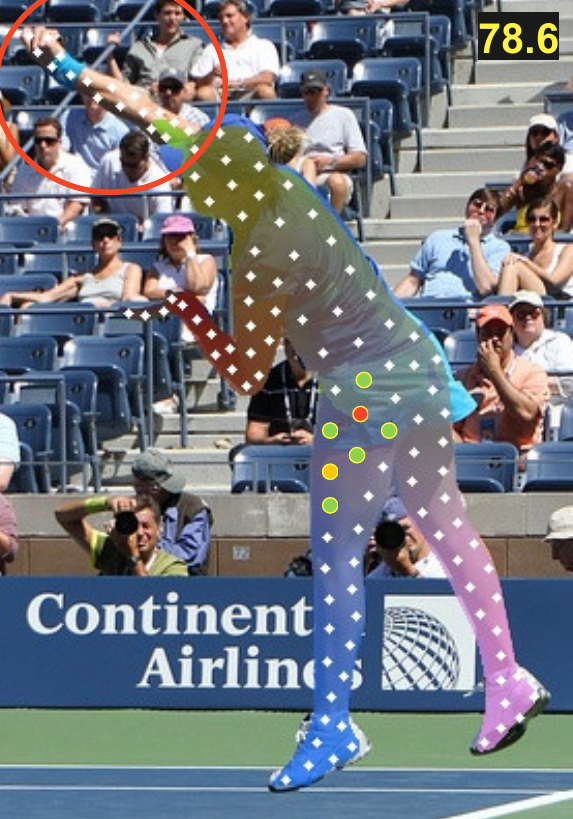}
    \end{subfigure}
    \hfill
    \begin{subfigure}{0.359\linewidth}
        \includegraphics[width=\linewidth,page=12]{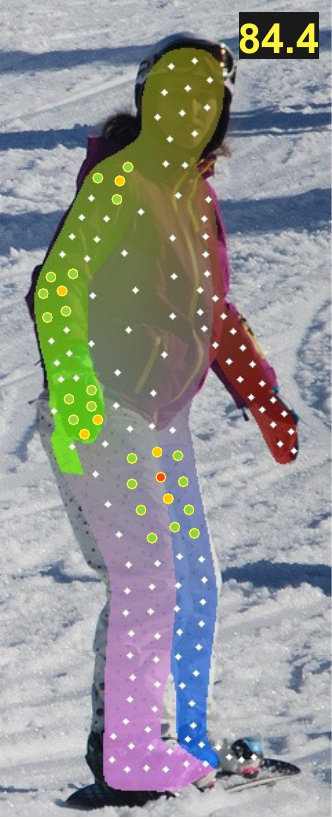}
    \end{subfigure}
    \caption{
    Images with GPS (geodesic point similarity) around 0.8. \colorbox{LightGray}{\textcolor{white}{Evaluation points}} are shown in white.
    Selected \textcolor{red}{wrongly estimated points} (similarity $<0.5$), \colorbox{LightGray}{\textcolor{yellow}{slightly wrong}} (similarity 0.5 – 0.9), and \textcolor{DarkGreen}{correct} (similarity $>0.9$).
    Typical errors are isolated wrong points among correct ones (left, hip), segmentation errors (left, red circle), and border points (right, legs).
    Loose clothing complicates annotation and estimation (right).
    }
    \label{fig:errors-analysis}
\end{figure}

In addition, we provide a detailed analysis of the variation in performance metrics for each evaluated sample. An example histogram, generated for \mbox{ViTPose-h wb}, is shown in \cref{fig:histogram}. We notice that the model maintains baseline precision on the vast majority of data samples and observe only a few performance drops, which are mainly caused by failure in the underlying pose inference. The worst are depicted in \cref{fig:errors-analysis}. However, these failures are largely compensated for by more common, sometimes drastic, improvements (shown in \cref{fig:examples}).

\section{Conclusions}
\label{sec:conclusions}

\begin{figure}
    \centering
    \begin{subfigure}{\linewidth}
        \includegraphics[width=0.32\linewidth]{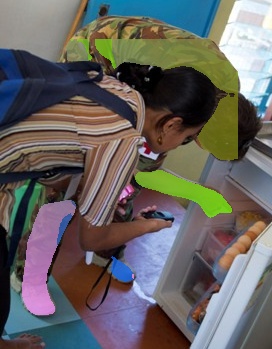}
        \hfill
        \includegraphics[width=0.32\linewidth]{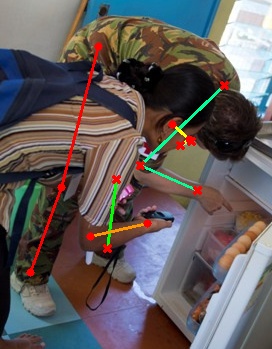}
        \hfill
        \includegraphics[width=0.32\linewidth]{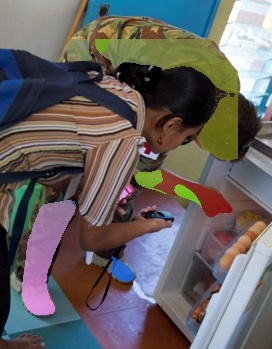}
    \end{subfigure}

    \begin{subfigure}{\linewidth}
        \includegraphics[width=0.32\linewidth]{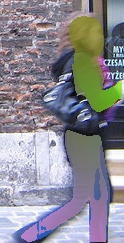}
        \hfill
        \includegraphics[width=0.32\linewidth]{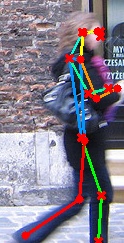}
        \hfill
        \includegraphics[width=0.32\linewidth]{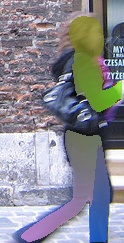}
    \end{subfigure}

    \begin{subfigure}{\linewidth}
        \includegraphics[width=0.32\linewidth]{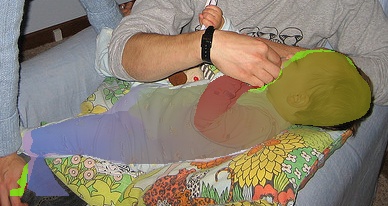}
        \hfill
        \includegraphics[width=0.32\linewidth]{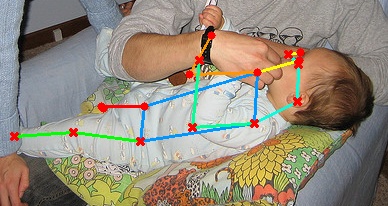}
        \hfill
        \includegraphics[width=0.32\linewidth]{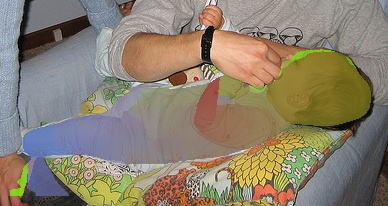}
    \end{subfigure}

    \caption{
    Three images with the largest performance decrease -- CSE (left), pose estimate (middle), PC-CSE (right). Pose conditioning reduces performance when the pose estimation fails. Despite the drop, the third most negatively affected image (bottom) shows only a 0.5\% decrease, highlighting that pose conditioning negatively impacts only a few images while improving many others.
    }
    \label{fig:pose-gone-wrong}
\end{figure}

We presented Pose-Constrained CSE (PC-CSE), a method that conditions UV map estimation using human pose.
PC-CSE leverages the robustness of 2D human pose estimation to provide global constraints, improving the consistency of UV map predictions produced by CSE \cite{DensePoseCSE}.

The original CSE \cite{DensePoseCSE} assigns pixels to vertices independently, which can lead to errors, such as assigning the same body part to multiple locations in the image and discontinuities in the same body part, as shown in \cref{fig:examples}.
\mbox{PC-CSE} introduces global supervision through pose constraints, ensuring that while pixel assignments remain independent, the global pose structure improves the consistency of the UV map.
This results in more coherent UV maps, free from artifacts and duplicated limbs.

Key findings are:

\begin{enumerate}
    \item Conditioning UV maps by pose, even with rudimentary constraints, provides consistent improvements, though overall performance gains remain modest.

    \item The choice of pose estimation model architecture has a negligible impact on the results.

    \item Whole-body skeletons enable more precise constraints for hands and feet, yielding small improvements over body-only skeletons without additional computational costs.

    \item COCO DensePose annotations are not entirely reliable; at least 1.5\% of the points are inconsistent with pose keypoints or are otherwise inaccurate. The accuracy of points under loose clothing remains uncertain as we could neither confirm nor disprove their precision.
\end{enumerate}

\textbf{Limitations.}
The primary limitation of \mbox{PC-CSE} lies in its reliance on precise pose estimation.
The method assumes that 2D human pose estimation (HPE) models are robust to challenges such as extreme poses, occlusions, and image deformations, which can condition UV map estimation effectively.
However, if the estimated pose is inaccurate, the constrained UV map will also be incorrect.
The most common errors occur in multi-body scenarios. 

Another limitation arises when two body parts are in close proximity.
For instance, when a person is sitting with crossed legs, pose constraints for both legs might overlap, preventing \mbox{PC-CSE} from correcting the original CSE estimates.
Although \mbox{PC-CSE} does not resolve such issues, it does not degrade overall performance.

\textbf{Future work.} The constraints implemented by us are very coarse, as they are satisfied by letting the pixel map \emph{somewhere} on the given body part. The corrections could become even more precise by taking the distance from its endpoints (keypoints) or the orientation of the body (frontal/dorsal) into account. In addition, there is substantial redundancy in the HPE and CSE representations, while the HPE algorithms are more advanced. The CSE method could be redesigned by building it on top of HPE and changing its objective to provide UV map estimation \emph{given a pose estimate} (and not just the image). We also plan to use the method for UV maps on animals using SMAL \cite{SMAL}.

\newpage

\textbf{Acknowledgements.} This work was supported by the Ministry of the Interior of the Czech Republic project No.~VJ02010041 and Czech Technical University student grant SGS23/173/OHK3/3T/13.

{
    \small
    \bibliographystyle{ieeenat_fullname}
    \bibliography{main}
}

\appendix
\maketitlesupplementary
\section{Annotation data quality assessment}
\label{sec:sup-quality}

In \cref{sec:quality}, we conduct research on the quality of annotations from the DensePose COCO dataset \cite{COCO,DensePose}. In order to efficiently identify the majority of erroneous annotations without having to manually examine the entire dataset, we take the following approach.

We establish several metrics to quantify the level of (in)consistency or ``(im)plausibility'' of each annotated example:
\begin{itemize}
    \item Proportion of the body segmentation covered by the instance mask. By definition, the body mask should be completely covered by the instance mask. This allows revealing problems 4 and 6, as enumerated in \cref{sec:quality}.
    \item Proportion of the area of the instance or body mask and the bounding box. Ideally, the mask should cover a significant portion of the bounding box. \\ This allows revealing problems 4, 5, and 6.
    \item Proportion of point-wise annotations within the instance or body mask. Human segmentation should ideally contain all (visible) keypoint annotations and dense correspondences, which concern the body, too. Likewise, this allows revealing problems 4, 5, and 6.
    \item Ratio of median points-to-bone distances. \\ We group ground-truth dense correspondences by body part and compute their median distance to the respective bone defined by ground-truth keypoint annotations (bone selection is done analogously to our mesh partitioning procedure, which exploits its resemblance to the COCO skeleton; see \cref{sec:method-UVbyPose}). We add up median distances for the same body part of either laterality. Then, we repeat the same procedure with the laterality of the keypoints flipped, and obtain another score. The ultimate value of the metric is the ratio of the two sums. When this value is high ($\gg 1$), it indicates a possibly confused laterality of keypoints or dense correspondences. \\ This allows revealing problems 1 and 3.
    \item Inference error. We run inference on all images from the dataset and compute the mean geodesic distance (error) per body part. High inference error usually indicates deficiencies in the model's performance, but, especially on training data, it might also help reveal annotation errors. We took advantage of repeated retraining and evaluation of the inference model (``human in the loop'') as it could initially have been overfitted to annotation errors. \\ This allows revealing problems 1 and 2.
\end{itemize}

We sort all annotations from the least consistent and manually examine them in this order until annotations with no apparent problems start to prevail.
This process is carried out individually for each defined metric.

\section{Ablation study on height estimation}
\label{sec:sup-height}

\begin{figure}
    \centering
    \begin{tikzpicture}
    \begin{axis}[
        width=\linewidth,
        height=0.5\linewidth,
        title={},
        xlabel={Frame},
        ylabel={Height (px)},
        xmin=-3, xmax=83,
        ymin=350, ymax=600,
        xtick={0, 10, 20, 30, 40, 50, 60, 70, 80},
        ytick={250, 300, 350, 400, 450, 500, 550},
        ymajorgrids=true,
        grid style=dashed,
    ]
    
    \addplot[
        color=red,
        line width=1pt,
    ]
    coordinates {
        (0.0, 444.94976479)
        (1.0, 450.07074201)
        (2.0, 447.96821648)
        (3.0, 437.15734508)
        (4.0, 439.30543816)
        (5.0, 441.37866737)
        (6.0, 441.96389756)
        (7.0, 459.23034159)
        (8.0, 456.02207607)
        (9.0, 453.00009456)
        (10.0, 449.27003345)
        (11.0, 444.22463844)
        (12.0, 450.94856197)
        (13.0, 451.35283201)
        (14.0, 456.42702455)
        (15.0, 449.87301427)
        (16.0, 444.01260492)
        (17.0, 454.70588122)
        (18.0, 458.55944301)
        (19.0, 436.41630155)
        (20.0, 427.44151869)
        (21.0, 436.98160853)
        (22.0, 447.51396048)
        (23.0, 438.77594855)
        (24.0, 469.91979182)
        (25.0, 445.71046335)
        (26.0, 471.1531992)
        (27.0, 465.69538483)
        (28.0, 451.69580056)
        (29.0, 435.89274978)
        (30.0, 449.18442004)
        (31.0, 444.28669141)
        (32.0, 421.16345249)
        (33.0, 441.76863527)
        (34.0, 447.9501352)
        (35.0, 449.49262884)
        (36.0, 436.6366571)
        (37.0, 428.42802785)
        (38.0, 425.92652395)
        (39.0, 425.12854774)
        (40.0, 432.70737791)
        (41.0, 420.26536491)
        (42.0, 421.87593137)
        (43.0, 424.16707872)
        (44.0, 418.90141408)
        (45.0, 427.33360256)
        (46.0, 423.61463176)
        (47.0, 419.5767928)
        (48.0, 429.90015756)
        (49.0, 427.1432392)
        (50.0, 419.23316669)
        (51.0, 430.24634433)
        (52.0, 428.94866787)
        (53.0, 423.04977305)
        (54.0, 414.42902043)
        (55.0, 407.74095265)
        (56.0, 427.60778872)
        (57.0, 444.87349911)
        (58.0, 466.83313864)
        (59.0, 460.42141008)
        (60.0, 473.57287649)
        (61.0, 475.27747773)
        (62.0, 439.79738262)
        (63.0, 452.68744848)
        (64.0, 470.19752667)
        (65.0, 462.16942353)
        (66.0, 429.96672902)
        (67.0, 420.51593813)
        (68.0, 475.22065889)
        (69.0, 426.94929359)
        (70.0, 437.51265751)
        (71.0, 407.50949571)
        (72.0, 423.40492895)
        (73.0, 434.74395923)
        (74.0, 404.29641026)
        (75.0, 405.07364813)
        (76.0, 395.55569391)
        (77.0, 429.65913552)
        (78.0, 390.95349159)
        (79.0, 430.73236592)
        (80.0, 431.29239355)
        };
    \addplot[
        color=blue,
        line width=1pt,
    ]
    coordinates {
        (0.0,438.438782131)
        (80.0,438.438782131)
    };
    \legend{Person height, mean}
        
    \end{axis}
    \end{tikzpicture}
    \caption{
    \textbf{Ablation on height estimation.}
    We infer pose from a dance video from \cite{TikTokDataset} at 10 frames per second and estimate the dancing person's height in pixels (red) using the algorithm in \cref{sec:method-areas}.
    The variable exhibits some noise due to pose changes, but remains within the interval of a few tens of pixels at all times.
    The bigger noise at the end of the video is caused by more extreme poses.
    }
    \label{fig:ablation-height}
\end{figure}

Our \mbox{PC-CSE} relies on estimating the proper outline of each constraint's region. In \cref{sec:method-areas}, we describe the algorithm we use to approximate the person's body measurements in pixel units of the image using only its inferred pose. The precision of such an algorithm can usually be determined by comparing the actual values and their algorithmic estimate on many images. We do not conduct such an experiment because of the lack of ground-truth data, but we verify its performance by taking a different approach.

The goal is to demonstrate that the estimate is not dramatically influenced by pose variations. However, images of people ``in the wild'' usually also differ in the distance of the person from the camera, as well as the underlying camera parameters. For a sensible comparison, these two factors need to remain constant. We notice that this requirement is met, for example, by short videos of people dancing in front of the camera uploaded to social networks such as TikTok \cite{tiktok}.

Therefore, we take advantage of the TikTokDataset \cite{TikTokDataset} and select several videos where a person performs a dance in front of a static camera without moving around the place. We run pose inference per video frame and record the height estimate. An example chart recording the progress of one video is shown in \cref{fig:ablation-height}. The variable does exhibit some noise, approximately on the scale of tens of pixels, which can be attributed to pose variations and noise in the pose estimation, but it remains centered on its mean value throughout. We note that the actual noise influencing the estimate of bone width ($\Delta{}$) is much smaller since the bone width is a small fraction of the person's height.

\end{document}